\documentclass[5p,times]{elsarticle}

\journal{Future Generation Computer Systems}

\usepackage{amssymb}
\usepackage[figuresright]{rotating}
\usepackage{graphicx}
\usepackage{comment}
\usepackage{xurl}
\usepackage{xcolor}
\usepackage{booktabs}  
\usepackage{tabularx}  
\usepackage{listings}
\usepackage{wrapfig}
\usepackage{rotating}  

\definecolor{codeblack}{rgb}{0,0,0}
\definecolor{codegreen}{rgb}{0,0.6,0}
\definecolor{codegray}{rgb}{0.5,0.5,0.5}
\definecolor{codepurple}{rgb}{0.58,0,0.82}
\definecolor{backcolour}{rgb}{0.95,0.95,0.92}

\lstdefinestyle{mystyle}{
    backgroundcolor=\color{white},   
    commentstyle=\color{codeblack},
    keywordstyle=\color{magenta},
    numberstyle=\tiny\color{codegray},
    stringstyle=\color{codeblack}, 
    basicstyle=\ttfamily\footnotesize,
    breakatwhitespace=false,         
    breaklines=true,                 
    captionpos=b,                    
    keepspaces=true,                 
    numbers=left,                    
    numbersep=5pt, 
    xleftmargin={0.25cm},
    showspaces=false,                
    showstringspaces=false,
    showtabs=false,                  
    tabsize=2,
    language=Python
}

\colorlet{punct}{red!60!black}
\definecolor{background}{HTML}{EEEEEE}
\definecolor{delim}{RGB}{20,105,176}
\colorlet{numb}{magenta!60!black}

\lstdefinelanguage{json}{
    basicstyle=\normalfont\ttfamily,
    numbers=none,
    numberstyle=\scriptsize,
    stepnumber=1,
    numbersep=8pt,
    showstringspaces=false,
    breaklines=true,
    frame=lines,
    backgroundcolor=\color{white},
    literate=
     *{0}{{{\color{numb}0}}}{1}
      {1}{{{\color{numb}1}}}{1}
      {2}{{{\color{numb}2}}}{1}
      {3}{{{\color{numb}3}}}{1}
      {4}{{{\color{numb}4}}}{1}
      {5}{{{\color{numb}5}}}{1}
      {6}{{{\color{numb}6}}}{1}
      {7}{{{\color{numb}7}}}{1}
      {8}{{{\color{numb}8}}}{1}
      {9}{{{\color{numb}9}}}{1}
      {:}{{{\color{punct}{:}}}}{1}
      {,}{{{\color{punct}{,}}}}{1}
      {\{}{{{\color{delim}{\{}}}}{1}
      {\}}{{{\color{delim}{\}}}}}{1}
      {[}{{{\color{delim}{[}}}}{1}
      {]}{{{\color{delim}{]}}}}{1},
}

\begin{document}
\begin{frontmatter}

\title{LLM4VV: Developing LLM-Driven Testsuite for Compiler Validation}

\author{Christian Munley}
\author{Aaron Jarmusch}
\author{Sunita Chandrasekaran}

\address{Department of Computer \& Information Sciences, University of Delaware, 19716, Newark, U.S.A.}

\begin{abstract}
Large language models (LLMs) are a new and powerful tool for a wide span of applications involving natural language and demonstrate impressive code generation abilities. The goal of this work is to automatically generate tests and use these tests to validate and verify compiler implementations of a directive-based parallel programming paradigm, OpenACC. To do so, in this paper, we explore the capabilities of state-of-the-art LLMs, including open-source LLMs - Meta's Codellama, Phind's fine-tuned version of Codellama, Deepseek's Deepseek Coder and closed-source LLMs - OpenAI's GPT-3.5-Turbo and GPT-4-Turbo. We further fine-tuned the open-source LLMs and GPT-3.5-Turbo using our own testsuite dataset along with using the OpenACC specification. We also explored these LLMs using various prompt engineering techniques that include code template, template with retrieval-augmented generation (RAG), one-shot example, one-shot with RAG, expressive prompt with code template and RAG. 
This paper highlights our findings from over 5000 tests generated via all the above mentioned methods. Our contributions include: (a) exploring the capabilities of the latest and relevant LLMs for code generation, (b) investigating fine-tuning and prompt methods, and (c) analyzing the outcome of LLMs generated tests including manually analysis of representative set of tests. We found the LLM Deepseek-Coder-33b-Instruct produced the most passing tests followed by GPT-4-Turbo.

\end{abstract}


\end{frontmatter}

\section{Introduction}
\label{sec:introduction}
Large language models (LLMs) are capable of understanding natural language as input and performing tasks with that understanding. LLMs are capable of an array of tasks, from text generation to sentiment classification, code generation, document summarization, and more. 
OpenAI's GPT-4 scored 90th percentile on the Uniform Bar Exam, 99th percentile on the verbal GRE~\cite{openai2023gpt4}, among several other impressive scores on prestigious academic exams. Similarly, Anthropic's Claude 2 achieved a 76.5\% score on the multiple choice section of the Bar exam, and over 90th percentile on GRE reading and writing exams ~\cite{claude2}. 
LLMs are pre-trained on large, unlabeled datasets with self-supervised learning and can be fine-tuned with domain-specific datasets for improved performance on specific tasks 
Code-specific LLMs are trained on a large corpus of open-source code, such as stack overflow and GitHub, enabling the LLM to write functional and complex code, such as shown in Github Copilot and Codeium~\cite{copilot, codeium}. 

With the vast capabilities of LLMs, the work presented in this paper explores the applicability of the most suitable LLMs and prompt engineering techniques for an optimal generation of a validation and verification testsuite for high-performance computing (HPC) compilers from a standard specification which is mainly natural language.
More specifically, the goal of this approach is to check for correctness of C/C++/Fortran compiler implementations of directive-based programming models, OpenMP~\cite{OpenMP} and OpenACC~\cite{openaccspec}. To the best of our knowledge, this paper is the first of its kind. 
  While this paper particularly focuses on the directive-based programming model, OpenACC, the approach can be adapted to the OpenMP model as well (this project is also work in progress).
The performance of LLMs in programming tasks such as interview-style questions has been explored over the past couple of years~\cite{zhong2023study}. 
 However, the application of the approach to the validation of compilers based on a standard specification has not been explored yet.

Briefly, OpenACC is a directive-based programming model, which targets x86 architectures, A64FX and accelerators such as GPUs. A standard specification defines OpenACC~\cite{openaccspec}. 
 Open-source compilers such as GCC~\cite{openaccgcc}, LLVM/Clacc~\cite{denny2018clacc}, and vendors such as HPE Cray (for Fortran), and NVIDIA interpret the specification to develop suitable compilers. 
 GCC, Clacc, and HPE Cray compilers target X86, NVIDIA and AMD accelerator architectures, while NVIDIA compilers target X86, NVIDIA GPUs and A64FX architectures. OpenACC offers varying levels of control over the execution of a program and the flow of data to users. The model allows the compiler to make optimization decisions or allows for customized distribution of data and execution. This paradigm is beneficial for domain scientists who do not want to re-implement potentially complex domain applications in a low-level parallel programming language such as CUDA~\cite{cudasdk}, OpenCL~\cite{opencl}, SYCL~\cite{alpay2020sycl}. OpenACC is a widely popular programming model spanning several domains including climate~\cite{sawyer2014towards,kim2021gpu,lapillonne2014using}, computational fluid dynamics~\cite{sathe2016accelerating}, nuclear physics~\cite{searles2019mpi+}, biophysics~\cite{stack2022openacc} among several other domains.

\subsection{Motivation}
\label{subsec:motivation}
With the popularity of the programming model, it is critical that the compiler implementations compiles and executes the code correctly. Why is this a challenge? 
Different compiler developers tend to interpret the specification differently often leading to misinterpretations and ambiguities. The complexity of the definition of the features is one of the several reasons. Even with considerable efforts to enhance compilers, they remain prone to issues, such as inaccuracy, like all other software~\cite{Compilerlerner,surveychen,acmyang,Sankar1989ANO,compilerle}. The other obvious issues include compilation failures or execution failures that need to be identified and addressed. Also each compiler implementation offers varying levels of support for OpenACC and the coverage of the specification by a compiler varies too, for example HPE/Cray supports only Fortran. 
Nevertheless, the user should not be burdened with the above challenges nor should they be expected to tailor the usage of the implementations for their applications. So, the expectation is that the implementations should be the same at the level of abstraction for the user.

These above challenges directly call for the need to validate the compiler implementations. This can be done so by writing functional, corner, regression and stress tests. This can be an involved and recurrent task. We have been developing a validation and verification (V\&V) testsuite~\cite{openaccaaron} for OpenACC just like we have been developing a testsuite for OpenMP ~\cite{OpenMPVV2022}.

The task of designing those tests can be quite challenging and cumbersome, and still may not always be correct or complete. 
There are many ways that the features of the base language, such as C, C++, or Fortran, and the parallel programming paradigm, such as OpenACC or OpenMP, can be used together, and it is humanely impossible to cover them all. 
Tests in both the OpenACC and OpenMP testsuites are manually created, requiring the teams to stay up-to-date with the tests' generation and maintenance. The team members evolve during the project period. There is a lag when team members leave the project and new members are trained to create quality tests. The lag can range from three weeks to three months or more at times - this observation is made based on Chandrasekaran's (co-author) vast experience training both undergrad and graduate students over a period of 10+ years to work on parallel computing projects. This can harm the project as we lose continuity and are behind on deliverable. The above hardships serve as a motivation for this work that explores the possibility of using LLMs for a V\&V testsuite that can help with complementing the efforts to write these tests manually.  

Developers can ``shift'' focus to designing complicated corner case tests and stress test compiler implementations. LLMs require oversight but can be handy if the effort is front-loaded. LLMs can also help adapt an existing test to new or updated features and perform well at translation between languages. There are many things that LLMs offer developers, though they must be used with human oversight and caution.

This paper makes the following contributions: 
\begin {itemize}
\item Provides LLM-generated tests to check for correctness of OpenACC compiler implementations
\item Explores a fine-tuning dataset for this task
\item Evaluates performance of LLMs, prompt engineering techniques, and fine-tuning on test generation for OpenACC 

\end{itemize}
This paper's generated testsuites, prompts, fine-tuning dataset, and specification JSON are available at our GitHub~\cite{chrisllm}.

\section{Overview of LLMs}
    \begin{figure}[!t]
\centering
\begin{lstlisting}[frame=single, numbers=left, language=Python]
User: Write a code in C to verify compiler implementation of the OpenACC specification of the acc delete routine. 
Assistant:
...
#include <stdio.h>
#include <stdlib.h>
#include <openacc.h>

int main() {
    const int N = 1000;
    float *a = (float*)malloc(N*sizeof(float));
    float *b = (float*)malloc(N*sizeof(float));

    // Initialize the arrays
    for (int i = 0; i < N; i++) {
        a[i] = i * 1.0f;
        b[i] = 0.0f;
    }

    #pragma acc data copy(a[0:N])
    {
        // Update 'a' array on the device
        #pragma acc parallel loop
        for (int i = 0; i < N; i++) {
            a[i] = a[i] + 1.0f;
        }

        // Delete the 'a' array from the device
        #pragma acc exit data delete(a[0:N])

        // Attempt to update 'a' array on the device again
        #pragma acc parallel loop
        for (int i = 0; i < N; i++) {
            a[i] = a[i] + 1.0f;
        }
    }

    // Check the result on the host
    for (int i = 0; i < N; i++) {
        if(a[i] != i + 1.0f) {
            printf("Validation failed at index %d! Host: %f\n", i, a[i]);
            free(a);
            free(b);
            return EXIT_FAILURE;
        }
    }
    printf("Validation passed!\n");

    free(a);
    free(b);

    return EXIT_SUCCESS;
}
\end{lstlisting}

\vspace{.5cm}

\caption{Example prompt and generated test, with no augmentation through context retrieval, one-shot example, or a code template. This test is a false pass, because it does not test the correct feature i.e. \textit{acc delete routine}, rather the \textit{exit data directive.}}
\label{fig:SimplePromptOutput}
\end{figure}

Large language models are based on the transformer architecture introduced in~\cite{vaswani2023attention}. These models are trained on a large corpus of data, generally gathered from the internet. LLMs are trained in two stages: pre-training, and fine-tuning. Pre-training uses a large and general data set, while fine-tuning requires a task-specific data set. Both stages involve optimizing the LLMs parameters by predicting the next word, or token, in the training data set, and updating the parameters, or the weights and biases of the architecture, according to the difference between the actual and predicted next token. 

Powerful LLMs consist of a large number of parameters, on the order of billions. Closed-source LLMs such as OpenAI's GPT-4 \cite{openai2023gpt4} and Anthropic's Claude \cite{claude}, are available through a web service or an API, but the parameters are not published. Open-source LLMs such as Meta AI's family of Codellama LLMs \cite{rozière2023code} and Stanford's Alpaca \cite{alpaca}, are publicly available for local use, meaning the parameters are published and available for download. Some organizations choose to provide an infrastructure for using the LLM locally, such as for Codellama, while many are available on the Huggingface Hub, a community-driven platform with over 120k LLMs~\cite{wolf2020huggingfaces}.

The input to an LLM is text, which must first be tokenized or split into chunks of characters depending on the type of application. An example tokenization algorithm would first convert every character to UTF-8 encoding, where every byte is a token. Then, iteratively combine the most common pairs of bytes, defining a new token, known as byte pair encoding.  ~\cite{sennrich2016neural, zouhar2023formal, tokenizationwebster}. The input text is referred to as a prompt. Once tokenized, the input consists of a set of tokens, from which the LLM creates embeddings, which represent the meaning of the tokens in a high-dimensional vector space. Next, the input embeddings are augmented with positional encodings, because the ordering is not captured in the data otherwise. The LLM uses the whole set of embeddings in the decoding process to produce output. The primary basic function of a language model is next token prediction - given a set of input data, a prediction for the next most probable token is made. This process is referred to as causal language modeling. Through causal language modeling, LLMs are capable of various specialized tasks, such as sentiment classification (e.g., rating how positive or negative a statement is), question answering, document summarization, and code generation~\cite{tamkin2021understanding}. 

\subsection{Prompt Engineering Techniques}
\label{subsec:prompt}

To achieve high-quality performance on complex tasks, specialized input, or prompts, are often required. The most simple prompt is one simply requesting the desired behavior, e.g. ``Write a code in C to validate compiler implementation of OpenACC parallel construct.'' would be a simple prompt. An example of the output from this type of prompt is shown in Figure~\ref{fig:SimplePromptOutput}.
The example test here passes, but it does not target the correct feature, the acc delete routine in this case, so it is a false pass.

The amount of detail in the prompt regarding the task itself is an important factor in creating an effective prompt. The LLM does not have any prior information about the task, so unless specifically instructed to act a certain way, there is no reason to expect certain behaviour. 
In this work, we compare the effectiveness of a simple, one-sentence prompt, versus a prompt using multiple sentences describing in detail the task and the requirements. 
We call the detailed prompt in this work an ``expressive prompt''.

\begin{table*}[t]
\centering
\small
\begin{tabularx}{\textwidth}{l|X|X|X|X}
\toprule
LLMs / Benchmark & HumanEval+ pass@1 & MBPP+ pass@1 & GSM8K & TruthfulQA \\
\midrule
Codelama-34B-Instruct & 43.9\textsuperscript{~\cite{liu2023is}*}  & 52.9\textsuperscript{~\cite{liu2023is}*} & 32.7\textsuperscript{~\cite{roziere2023code}*} & 47.4\textsuperscript{~\cite{roziere2023code}} \\
Phind-Codellama-34b-v2 & 67.1\textsuperscript{~\cite{liu2023is}} & - & - & - \\
Deepseek-Coder-33b-Instruct & 75.0\textsuperscript{~\cite{liu2023is}} & 66.7\textsuperscript{~\cite{liu2023is}} & 60.7\textsuperscript{~\cite{guo2024deepseekcoder}*} & - \\
GPT-3.5(-Turbo) & 70.7 (turbo)\textsuperscript{~\cite{liu2023is}} & 69.7 (turbo)\textsuperscript{~\cite{liu2023is}} & 57.1\textsuperscript{~\cite{openai2023gpt4}} & 47.0\textsuperscript{~\cite{openai2023gpt4}} \\
GPT-4(-Turbo) & \textbf{81.7} (turbo)\textsuperscript{~\cite{liu2023is}} & \textbf{70.7} (turbo)\textsuperscript{~\cite{liu2023is}} & \textbf{92.0}\textsuperscript{~\cite{openai2023gpt4}} & \textbf{60.0}\textsuperscript{~\cite{openai2023gpt4}} \\
\bottomrule
\end{tabularx}
\caption{Performance of LLMs using relevant benchmarks in percentages. The metric is pass@k metric that indicates unit test pass rate when selecting respectively k samples from the candidate solutions \cite{chen2021evaluating}. The pass@k metric for GSM8k and TruthfulQA seemed unclear. We could not find the results for all LLMs and benchmarks. We have cited references for each value. Results for some of these LLMs are for the base LLM as indicated in superscript~\textsuperscript{*}.} 
\label{Tab:performance}
\end{table*}

LLMs are capable of learning to complete a new task based on a few examples provided in the prompt, known as few shot prompting \cite{brown2020language}. For example, if the task is translation, the prompt could include a few example translations before requesting translation of another sentence. This technique can be limited to a single example, known as one-shot prompting. If the task is code generation for interview-style programming questions, a good solution to an example question would be provided as a one-shot example. Alternatively, providing no example of the desired behavior in the prompt is called zero-shot prompting. Few or one-shot prompting can teach the LLM the desired behavior from the prompt, rather than from specific training. This enables a single, general model to be applied to various specialized tasks without additional training for each task. Even a single example can provide the model with information on how to perform the task properly.

Retrieval-augmented generation (RAG) is a common technique to engineer effective prompts by including relevant context to the task, retrieved with a search algorithm, in the prompt \cite{lewis2021retrievalaugmented}. Typically, this is implemented by splitting a text database into chunks, creating embeddings from the chunks that represent the meaning, and storing the embeddings in a vector database. Then, the vector database can be searched for similar information to the prompt, and the retrieved data can be wrapped into the prompt with a template.  For example, in a domain-specific question answering task, Wikipedia pages from the domain can be converted into a vector database, and queried for relevant information to each question, to be included in the prompt as context.

Other strategies exist to improve LLM output through prompt engineering, such as chain-of-thought prompting, in which the LLM is instructed to plan its solution to the task, before completing the task \cite{wei2023chainofthought}. This method can enable the LLM to tackle more complex tasks by planning them out before performing them. A more advanced approach involves creating a parameterized prompt that can be optimized to produce the desired output through training, known as prompt tuning \cite{lester2021power}. Prompt engineering is new and broad area, so the optimal methods are still being learned and there are no gold standards that indicate which prompt would be an idea choice for a given case study.

For our work, we explore expressive prompting, one-shot prompting, and RAG. We choose to evaluate these methods for a few reasons. First, they have not previously been evaluated on this task. Next, all three methods are used in order to improve the quality of test generation in different ways. The goal of using expressive prompts in this research is to evaluate the effect of the length of the task description in the prompt on the quality of output. Creating a validation test for OpenACC features is not a simple task, so an expressive prompt seems necessary. We evaluate one-shot prompting as a method that attempts to teach the LLM the proper style of testing OpenACC features, without additional fine-tuning. Finally, we use RAG in this research because its crucial that the LLM is aware of the latest OpenACC specification to create accurate tests, so it must be included either through training or RAG. 

\subsection{Fine-tuning of LLMs}

    Prompt engineering is a powerful method to adapt a model to specific tasks without requiring expensive retraining. 
    While prompts are useful, they do not alter the underlying parameters of the language model, thus constraining their ability to specialize or improve performance. 
    Fine-tuning is training a foundation model on a domain-specific dataset, which involves updating all of the model's parameters. 
    This method can teach a LLM to solve a new task without examples in the prompt \cite{wei2022finetuned}. Fine-tuning can also be more robust than prompt design, as the model can be shown more examples than can fit in the prompt. Additionally, updating the LLMs parameters during fine-tuning adapts the model to a specific task, while relying on a prompt leaves the LLMs architecture unchanged.  
    The maximum input size to a LLM, sometimes called the context window, is limited, while the training datasets can be enormous. LLMs can't see all of the fine-tuning data during inference, they can only operate based on their parameters and input. 
    Using prompts offer the advantage of making all the data directly available to the language model. Though fine-tuning can be costly, parameter-efficient methods can reduce costs, such as freezing the model weights during fine-tuning and introducing small trainable layers into the architecture~\cite{vonwerra2022trl, dettmers2023qlora}.

    LLMs are trained on a large corpus of data from the internet, but that does not necessarily include a lot of information about the subject of our interest, i.e., OpenACC. 
    In order for the LLMs to correctly use and validate OpenACC implementations, they need to learn the features of OpenACC and their definitions. This can be achieved through context within the prompt or through fine-tuning on an OpenACC-specific dataset.

\subsection{Reinforcement-Learning with Human Feedback (RLHF)}
\label{subsec:rlhf}
    Pre-training and fine-tuning are suited to teach a model next token prediction or other similar objectives. However, defining the quality of generated text (or a validation test) is complex, and difficult to capture in a loss function based on something like next token prediction. 
    RLHF ~\cite{huggingface-blog, NIPS2017_d5e2c0ad} enables developers to train LLMs based on human preference, i.e. based on labeling of outputs as desirable or not by a human. This is the training technique that enabled ChatGPT~\cite{ouyang2022training} to go from a text generation tool to a friendly assistant who usually refuses to help you with unethical tasks. In generating compiler validation tests for OpenACC, this may be a useful technique, as the quality of a test is not clearly defined by a metric such as pass/fail, so human feedback may be an important ingredient.

\subsection{LLM Benchmarks}
\label{subsec:benchmark}
    Performance of LLMs is evaluated on a variety of benchmarks, often task-specific, such as shown in Table~\ref{Tab:performance}.
    This table focuses on results from the selected LLMs that we have used for our research. 
    Currently to the best of our knowledge, there is no gold standard for test generation using LLMs or via natural language processing. Manually generating tests have so far been the gold standard for this task. To achieve such a standard for auto test generation, we need to leverage multiple ideas. Ideas could include either training our own LLM or using an existing LLM for the task. Since the former is a time and energy consuming task, we are currently pursuing the latter idea i.e. explore existing LLMs for its suitability to auto-generate tests. In order to do so there are two important steps. The first step is to choose the LLM that is the most suited for our task. The second step is to prepare this LLM for the task. 
    
    With respect to step one, i.e. choosing the right LLM, we are not aware of an LLM benchmark that is particularly trained on programming tasks and languages specifically. The closest we found was HumanEval that is an evaluation set released by OpenAI to evaluate functional correctness of Python code generation from docstrings~\cite{chen2021evaluating}.  The HumanEval benchmark is relevant to this work as it evaluates code generation performance from natural language specification at varying levels of complexity.  Although the benchmark's scope does not directly cover test generation quality and feature coverage, it is relevant to the evaluation of code generation from natural language prompts. 
  The second best we found was MBPP benchmarks ~\cite{austin2021program}. MBPP is similar to HumanEval, designed to evaluate the performance of LLMs on python programming tasks suitable for entry-level programmers. 
      
    To inform our selection on relevant skills beyond programming, we considered two additional benchmarks. GSM8K is a benchmark of 8.5K grade school math word problems, introduced by OpenAI in~\cite{cobbe2021training}, designed to evaluate LLM performance in multi-step mathematical reasoning. We find this benchmark suitable for this research because writing validation and verification tests from the standard specification requires multi-step reasoning and mathematical validation schemes. We also chose TruthfulQA, comprised of 817 questions spanning various categories designed to evaluate LLM capability of avoiding common misconceptions and truthfully answer questions.~\cite{lin2022truthfulqa} We selected this benchmark as we need the LLM to accurately interpret and use the OpenACC specification, rather than succumb to common misconceptions.

With respect to step two, the chosen LLM needs to be trained to learn from a variety of inputs. These inputs include hundreds of manual written tests, the standard specification of the programming model under consideration. After training, a RAG-based approach can further improve the quality of generative AI by offering context-rich and information dense outputs. 

Together, via an iterative and analytic process, there is potential to create a gold standard for auto-generation of tests.

\begin{figure}[!h]
\centering
\begin{lstlisting}[frame=single, numbers=left, language=Python]
f'''Write a code in {language} to verify compiler implementation of the OpenACC specification of {feature}. 

Make sure to follow the template of the format provided. Include the provided header file, and any other necessary libraries.
Write simple code to test {feature} in {language}. Try to isolate that feature while still using it correctly.
This code is part of a testsuite that will be ran, so write complete code, don't leave it unfinished.
The goal is to return 0 if the target feature, {feature}, is working properly, and not zero otherwise.
The context below is from the most recent OpenACC specification, make sure to refer to it to produce up to date code.

Context: {context}

Template: {template}
'''
\end{lstlisting}
\caption{Expressive prompt with RAG and Template. This prompt includes detailed instructions on the task, and provides context from the OpenACC specification with a code template mentioned below the context.}
\label{fig:DetailedPrompt}
\end{figure}

    \section{Related Work}
\label{sec:relatedwork}
    We will start off this section by sharing our own on-going and relevant OpenACC V\&V testsuite project~\cite{openaccaaron} where we have been manually generating functional tests. 
    This work stands as a good point of reference for the development of LLM work discussed in the rest of the paper. Quite similar to the OpenACC V\&V is our other on-going project on OpenMP offloading V\&V~\cite{OpenMPVV2022} which also serves the similar purpose of manually generating functional tests. While this current work focuses on OpenACC V\&V, work is in progress to adapt lessons learnt for OpenMP V\&V as well. 
    
    We next summarize some of the relevant and most recent work in this area of research. Many LLMs have been developed for code generation. Meta released a family of Codellama LLMs, OpenAI released Codex and GPT-4, among others, Github created Copilot \cite{copilot}, Amazon created Codewhisperer \cite{codewhisperer}, BigCode created StarCoder \cite{li2023starcoder}, Microsoft created WizardCoder \cite{luo2023wizardcoder}, to name some. Fine-tuned code generation LLMs from foundation models are also prevalent in the open-source community, such as Phind's fine-tuned LLMs.
    
    Some related works have tried other approaches to improve LLM performance for high-performance computing tasks. In LM4HPC~\cite{chen2023lm4hpc}, Chen et. al. create a HPC-specific tokenizer, designed to more accurately split the code into tokens corresponding to the typical syntax of HPC code. Moreover, LM4HPC presents multiple HPC-specific training datasets. These are useful both in constructing large pre-training datasets or for domain-specific fine-tuning. LM4HPC also develops three pipelines, code similarity analysis, parallelism detection, and OpenMP question answering, and creates leaderboards from evaluation of performance on these tasks. 
    
    Kadosh et al.~\cite{kadosh2023scope} present Tokompiler, also a tokenizer designed for HPC tasks.
    Godoy et al.~\cite{Godoy_2023} evaluates LLM performance on generation of HPC numerical kernels using GitHub Copilot powered by OpenAI's Codex.  
    Building upon this work, the authors in~\cite{valero2023comparing} evaluate the performance of Llama2 and GPT-3 LLMs for HPC kernels generation. 
   The authors in ~\cite{nichols2023modeling} explore fine-tuning LLMs with HPC-specific data and testing on downstream tasks such as code completion, OpenMP labeling, and performance prediction.


    At Microsoft,  LLMs are used to create CodeT~\cite{chen2022codet}, a method that involves first generation of test cases for code solutions to programming tasks, then execution of the code samples using the generated tests, and evaluation of output. Tufano et al.\cite{tufano2021unit} at Microsoft present AthenaTest, an approach using the BART transformer to generate unit tests. Shafer et al.~\cite{schäfer2023empirical} present a approach, TestPilot, that enables the LLM to attempt to fix failing tests .

        LLMs are being gradually adopted for HPC problems. While the literature review refers to test generations using LLMs, they are not necessarily compiler validation tests, but provide a similar workflow. Our research stands apart from these projects in that it evaluates the performance of LLMs on a) comprehending a lengthy natural language programming paradigm specification, and b) subsequently generating validation tests for compiler implementations. This paper, to the best of our knowledge, is the first in this direction.

\section{Methods}
This section provides a detailed description of how we used LLMs, prompt engineering techniques and datasets to automatically generate and evaluate tests for OpenACC.
Our choices were grounded in recent benchmarks and related work.

\begin{figure*}[!t]
\begin{lstlisting}[frame=single, numbers=left, language=json]
{"prompt": "<prompt text>", "completion": "<ideal generated test>"}
{"prompt": "<prompt text>", "completion": "<ideal generated test>"}
\end{lstlisting}
\caption{Format of LLM fine-tuning datasets that shows a set of prompts and desired responses}
\label{fig:FinetuneFormat}
\end{figure*}

\subsection{Selection and initial expectations from LLMs}	
   In this subsection, we somewhat set the stage on what to expect out of the LLMs that we would use for our work so that we can eventually decipher how close or far we were from our expectations. Based on the benchmarks described in Section~\ref{subsec:benchmark} and performance tabulated in Table~\ref{Tab:performance}, we selected the following LLMs for this research: OpenAI's GPT-3.5-Turbo and GPT-4-Turbo, Meta AI's Codellama-34b-Instruct, Phind's fine-tuned Phind-Codellama-34b-v2, and Deepseek's Deepseek-Coder-33b-Instruct~\cite{brown2020language, openai2023gpt4, rozière2023code, Phind_2023, guo2024deepseekcoder}. 
    There are no verified sources for OpenAI that we found that provides information on the size of GPT-4, in other words, the number of parameters. 
    The largest version of GPT-3 is 175B parameters, much larger than the open-sourced LLMs. As GPT-4 is GPT-3's successor, GPT-4-Turbo is presumably larger than any other model and scores high on a variety of benchmarks, so we initially expect the best performance from this LLM. 
    However, Deepseek-Coder-33b-Instruct scores competively against GPT-4-Turbo on the HumanEval+ benchmark as shown in Table~\ref{Tab:performance}. Additionally, Codellama and Deepseek LLMs were trained specifically for code generation, while GPT-4 is general purpose, so we expect all of the open-source LLMs to be competitive.

\subsection{Prompt creation}
\label{sec:prompt}

    To create V\&V tests for validation of OpenACC compilers using LLMs, we create sets of prompts requesting a test for each feature of OpenACC as listed in Chapters 2 and 3 of OpenACC specification's table of contents \cite{openaccspec}. Please refer to Section~\ref{subsec:prompt} for an overview of prompt engineering techniques. 

    To explore the effectiveness of an expressive prompt, retrieval-augmented generation (RAG) and one-shot prompting methods, we create multiple sets of prompts both incorporating and excluding these methods. For all tests without one-shot examples, we choose to provide a code template in the prompt, to instruct the LLMs to produce standardized output for testing and evaluation.
   Thus, we create the 5 following sets of prompt methods and in the upcoming subsections, we narrate them in detail in the following subsections: 
   \begin{itemize} 
    \item Template 
    \item Template + RAG
    \item One-shot 
    \item One-shot + RAG
    \item Expressive + Template + RAG 
\end{itemize}

\subsection{Template}
A 'Template' is defined in this work is defined as a manual test from our previous work~\cite{openaccaaron}, with the testing logic removed. We provide a template for the LLM to format the test, so that the test format is unified and works within our testsuite infrastructure.

\subsection{Retrieval-Augmented Generation (RAG)}
    To provide relevant context from the OpenACC specification in each prompt, we implement RAG using two methods: a similarity search algorithm, and manual retrieval of context with a JSON of the specification. 
 Providing the most recent OpenACC specification to the LLMs is essential to testsuite generation because we do not know that the LLMs are pre-trained on the latest version of the specification, if any. Moreover, if pre-trained on the latest specification, the LLMs can only access information from the specification through the parameters of the model which are optimized during training. The training data is not perfectly stored in the parameters, leading to LLMs producing false facts, known as hallucinations~\cite{mckenna2023sources}. 
 
 To enable accurate factual recall, RAG methods are implemented to provide the LLMs with direct access to relevant facts during inference. 
 In this work, we provide relevant pieces of the specification in the prompt to the LLMs to evaluate the effectiveness of RAG for improving LLMs ability to recall the specification accurately, rather than improperly use OpenACC in the generated test.

To manually retrieve context, we first construct a JSON of the specification with the table of contents as the keys and the corresponding sections as values. This is available in our GitHub~\cite{chrisllm}.
For each prompt, we include the corresponding section in the prompt using the JSON. We believe that the JSON would also be valuable to the OpenACC community for projects like this. 
   
    To perform similarity search retrieval, we create a vector store containing the specification text. 
    First, we split the text into chunks of 1000 characters, with an overlap of 100 characters - this would give some continuity between chunks. We then create embeddings of the chunks that represent their meaning~\cite{app12178805}.
    Next, we perform a vector store similarity search for each prompt to retrieve relevant information from the specification to the prompt. The retrieved information is then wrapped into the prompt, shown in Figure~\ref{fig:DetailedPrompt}. 
    The necessity of context retrieval arises because the context window of most LLMs is shorter than the length of the entire specification. 
    The whole specification can not fit into the prompt, otherwise an alternative approach would be to include the entire specification in the prompt for each test. 
    
   Some LLMs offer a context window that would fit the entire specification, but by providing only relevant information reduces the amount of data it must filter. 
For example, Meta AI trains all Codellama LLMs with a context window of up to 100k tokens, which is longer than the specification. 
    This means we are able to fit the entire text into the prompt instead of performing RAG. 
   We did so with Codellama-34b-Instruct using Meta AI's published inference code~\cite{codellamagit} which only supports parallelization of four GPUs. We observed an out-of-memory issue. Had Meta supported more than four GPUs, we would have overcome this challenge. We are not aware if Meta plans to support more than four GPUs in the near future. 
    We also tried Codellama using Huggingface transformers API~\cite{wolf2020huggingfaces}, which supports parallelism with more than four GPUs, however we observed incorrect output so we tabled it for near-future work.

\subsection{One-shot vs Zero-shot Prompting}	
    We compare the performances of zero-shot prompting vs. one-shot prompting by creating two set of prompts, one for each.
    To construct zero-shot prompts, we do not provide any example of validating OpenACC compiler implementation, however we choose to include a code template to help standardize the generated tests across testsuites. 
    For the one-shot prompts, we provide within each prompt an example prompt and a correct manually written OpenACC V\&V test from the OpenACC V\&V testsuite~\cite{openaccaaron}. 
    We run both sets of prompts with all selected LLMs. 
    Additionally, we evaluate the performance of one-shot prompting with RAG.

    \subsection{Expressive Prompt}
    
Besides training, the prompt is the only input that a LLM uses to produce the desired output. The prompt is highly influential on the output of the LLM. Even the number of words, choice of words (often termed as word-choice)  impacts the quality of the output. 
In our work we compare the output from a simple prompt with one sentence, versus a detailed prompt with various requirements listed for the task. 
To this end, we compare the output between the methods that create the most passing tests without an expressive prompt, to the same methods using an expressive prompt, shown in Figure~\ref{fig:DetailedPrompt}. Improved output with expressive prompt using RAG with code template is shown in Figure~\ref{fig:DetailedPromptOutput} (in contrast with a simple prompt and output, shown in Figure~\ref{fig:SimplePromptOutput}).

    \subsection{Dataset and Fine-tuning}
    \label{sec:finetuning}

    We highlight the steps taken to construct a fine-tuning dataset using our manually created OpenACC V\&V testsuite and the OpenACC specification and perform fine-tuning of LLMs to improve quality of validation test generation for OpenACC. 
    Each manually written test focuses on one feature, so we create a prompt for each test. Each prompt in the fine-tuning datasets requests a validation test for OpenACC implementations of the feature being tested. The format of LLM fine-tuning datasets is typically a set of prompts and desired responses, as shown in Figure~\ref{fig:FinetuneFormat}.

    For each manually written test, we find the most relevant section from the OpenACC specification to use as context in the prompt. In this way, we implement RAG in the fine-tuning dataset, providing two benefits. First, it resembles the prompting method used after training for inference. Second, by fine-tuning on the specification along with manually written tests, the LLMs gain exposure to both the specification and properly written validation tests. We create one dataset comprising of 1335 examples in C, C++, and Fortran, as described above.
    One-shot examples and templates are not included in the fine-tuning dataset because the goal of fine-tuning is to teach the LLM the desired behaviour without providing instruction in the prompt.

    Throughout fine-tuning development, we evaluate the LLM output to ensure that it is tailored to OpenACC and producing relevant code. 
    The LLMs we fine-tune in this research are GPT-3.5-Turbo, Phind-Codellama-34b-v2, and Deepseek-Coder-33b-Instruct. 
    GPT-4-Turbo is not available publicly for fine-tuning at this time, so we include GPT-3.5-Turbo for this purpose. We report results in Section~\ref{sec:results}.

\begin{table*}[t]
\small  
\begin{tabular*}{\textwidth}{@{\extracolsep{\fill}}lccccc}
\toprule
LLMs / Methods & Template & Template + RAG & Oneshot & Oneshot + RAG & Expressive + Template + RAG \\
\midrule
Codellama-34b-Instruct & 15 & 11  & 7  & 4  & 35  \\
Phind-Codellama-34b-v2 & 48  & 43  & 52  & 22  & 43  \\
Deepseek-Coder-33b-Instruct & \textbf{51 } & \textbf{56 } & 51  & \textbf{57 } & 47  \\
GPT-3.5-Turbo & 15  & 19  & 5  & 11  & 23  \\
GPT-4-Turbo & 48  & 45  & \textbf{54 } & 49  & \textbf{54 } \\
Fine-tuned GPT-3.5-Turbo &  &  &  &  & 27  \\
Fine-tuned Phind &  &  &  &  & 44  \\
Fine-tuned Deepseek &  &  &  &  & 47  \\
\bottomrule
\end{tabular*}
\caption{Stage 1 results displaying passing percentage for all selected LLMs and five methods. We see that Deepseek-Coder-33b-Instruct generally produced the most passing tests, though in some cases GPT-4-Turbo produced more - these are indicated \textbf{in bold}. We do not have results for the Fine-tuned GPT-3.5-Turbo, Phind and Deepseek LLMs because our focus is on exploring the Fine-tuning method of these LLMs against prompt-based methods rather than "with" the prompt-based methods. However we did run the three Fine-tuned LLMs with Expressive + Template + RAG prompts as we found "with" this prompt-based methods, the non-Fine-tuned LLMs achieved the best performance.}
\label{tab:stage1}
\end{table*}


\begin{figure}[!b]
\centering
\begin{lstlisting}[frame=single, numbers=left, language=Python]
#include "acc_testsuite.h"

#ifndef T1
/* T1: acc delete, V:2.0-2.7 */
int test1() {
    int err = 0;
    int* h_data;
    size_t bytes = sizeof(int);

    h_data = (int*)malloc(bytes);
    *h_data = 15;

    /* Place data on device */
    acc_copyin((void*)h_data, bytes);

    /* Delete data from device */
    acc_delete((void*)h_data, bytes);

    /* Data should now not be present in the device */
    if (acc_is_present((void*)h_data, bytes)) {
        err = 1;
    }

    free(h_data);

    return err;
}
#endif

int main() {
    int failcode = 0;
    int failed;

#ifndef T1
    failed = 0;
    for (int x = 0; x < NUM_TEST_CALLS; ++x) {
        failed = failed + test1();
    }
    if (failed != 0) {
        failcode = failcode + (1 << 0);
    }
#endif

    return failcode;
}
\end{lstlisting}
\caption{Generated test with Expressive + Template + RAG prompt. This is a passing test, targeting the correct directive, and using the desired format for the testsuite infrastructure and is generated based on context from the specification.}
\label{fig:DetailedPromptOutput}
\end{figure}

\subsection{Types of errors}
   Before we talk about the process of development, let us categorize the different types of errors and specify the intended outcome. A test is labeled ``pass'' if it returns 0, indicating successful outcome. Alternatively, there are three potential types of errors: 
   \begin{itemize}
       \item   A parsing error occurs when the generated output is incomplete, and no end to the code block is signified. This occurs when the LLM gets stuck in a infinite loop of generation, instead of generating an end to the test. The script we use to parse the output looks for an end to the code, through back-ticks or a return statement - a parsing error occurs when neither appears. 
       \item A test is labeled compile error if it does not successfully compile.
       \item A test is labeled runtime error if it fails during execution or returns nonzero. 

   \end{itemize}
   
    The prompts and fine-tuning methods we implement are designed to generate tests that return a value of 0 if the feature is working correctly and a non-zero value otherwise. 
    The output of LLMs is typically stochastic, so it cannot be guaranteed that the tests that are generated indeed return 0 only if the feature is working correctly. 
    It may be the case that the LLM writes a test that returns 0 regardless, so the tests require further evaluation.
    

\subsection{Three Stages}
\label{sec:stages}

    Here we further explain the development process. We have broken down the process into three stages for easier comprehension. We further explain what entails in each of these stages. 
    
    \textbf{Stage 1:} First, we use each selected LLM and all 5 methods described in Section~\ref{sec:prompt} to generate 95 tests. These are only written in C. These tests cover every OpenACC features listed in chapters 2 and 3 of the specification. We record the results of these generated tests against one OpenACC compiler.
    
    We do not create permutations of tests in this stage, such as creating a test for each clause for each compute construct, i.e. parallel, serial, and kernels. 
    The script we use to compile and run the generated tests labels each test as either parsing error, compile error, run-time fail and pass (definitions can be found in the previous section). Table 2 presents the passing percentage for every method we have used for LLMs and their fine tuned versions. We observed that GPT-4-Turbo and Deepseek-Coder-33b-Instruct produced the most passing tests.
  
  
    While we do not collect statistics on expert analysis of the tests at this stage, we manually assess the generated tests and consider the results of the runtime scripts in selecting the best approach for Stage 2 which entails further refining our development process.

   \textbf{Stage 2:} 
    Based on what we observe in Stage 1, we drop the following methods: code template, RAG with code template, one-shot example, RAG with one-shot example, retaining the expressive prompt using RAG with code template in this stage. For each LLM, including a fine-tuned version of each where applicable, we generate 351 tests for every OpenACC feature in C, C++, and Fortran listed in Chapter 2 and 3 of the specification.  Additionally, we generate permutations of the compute construct clause tests for each compute construct, i.e. parallel, serial, and kernels, to increase coverage of the specification. Again we record the results of running each generated test against an OpenACC compiler, and select the LLM with the largest number of passing tests for manual analysis in Stage 3. Table~\ref{tab:stage2setup} lists the full experimental setup for Stage 2 generation. We were not able to pin down the number of GPUs that GPT-4-Turbo and the fine-tuned GPT-3.5-Turbo LLMs used. Table~\ref{tab:stage2results} displays all results for all the LLMs chosen. 
    Table~\ref{tab:passperlang} shows the passing percentage per base language. In comparison to the runtime displayed in Table~\ref{tab:stage2setup}, we benchmarked Deepseek-33b-Coder-Instruct running on 1x AMD EPYC 7763 CPU, and found the same testsuite generation would take approximately 190 hours, thus using 4 x NVIDIA A100s GPUs provided a 52x speedup, by this evaluation. Note that we did not use significant optimizations for CPU or make any special efforts to fully optimize GPU utilization.

\begin{table*}[!thb]  
\centering

\small  
\begin{tabularx}{\textwidth}{l|XXXXX}
\toprule
LLMs & 
Tests generated & 
GPU Time Taken & 
GPUs used & 
Model Parameter \\ 
\midrule
Codellama-34B-Instruct & 351 & $\sim$3.6 hours & 4 A100s & 34B  \\ 
Phind-Codellama-34B-v2 & 351  & $\sim$3.6 hours   & 4 A100s  & 34B \\ 
Deepseek-Coder-33b-Instruct & 351  & $\sim$3.6 hours   & 4 A100s  & 33B \\ 
GPT-4-Turbo  & 351 & $\sim$4 hours & Unknown & Unknown \\ 
Fine-tuned GPT-3.5-Turbo & 351 & $\sim$4 hours & Unknown & 174B \\ 
Fine-tuned Phind-Codellama-34B-v2 & 351  & $\sim$3.6 hours   & 4 A100s  & 34B \\ 
Fine-tuned Deepseek-Coder-33b-Instruct & 351  & $\sim$3.6 hours   & 4 A100s  & 34B \\ 
\bottomrule
\end{tabularx}
\caption{Experimental setup for Stage 2 using the LLMs, including the number of tests generated, GPUs used and the time they took for inference}
\label{tab:stage2setup}
\end{table*}

\begin{table*}[!tb]
  \centering

    \begin{tabular}{l|cccc}
    \toprule
    LLMs & Parsing Error & Compile Fail & Runtime Fail & Pass \\
    \midrule
    Phind-Codellama-34b-v2 & 3     & 158 & 71 & 119 \\
    Codellama-34b-Instruct & 66    & 216   & 22    & 47 \\
    Deepseek-Coder-33b-Instruct & 17     & 105     & 59     & \textbf{170} \\
    GPT-4-Turbo & 11     & 127   & 71    & 142 \\
    Finetuned GPT-3.5-Turbo  & 17     & 186   & 63  & 85 \\
    Finetuned Phind & 109 & 80 & 69 & 93 \\
    Finetuned Deepseek & 18 & 108 & 65 & 160 \\
    \bottomrule
    \end{tabular}%
  \caption{Stage 2 results (all 
 base languages (C/C++ \& Fortran), permuted constructs and clauses, Expressive + Template + RAG). We note that Deepseek-Coder-33b-Instruct produced the most passing tests (the number of tests generated, \textbf{in bold}, though it requires a strict manual evaluation of validation performance of OpenACC usage done in Stage 3}
  \label{tab:stage2results}%
\end{table*}%

   \textbf{Stage 3:} We produced 35 testsuites in this work, generated from a total of 5,117 prompts, thus the entire output is too large to assess by hand. Here we manually analyze the output of the LLMs and methods that generated the most passing tests from Stage 2. Through this manual analysis, we aim to determine where the LLMs and methods fall short in generating correct tests and consider how the quality of test output may be improved. We manually analyze a representative subset of both passing and failing tests to determine whether the OpenACC implementations are correct. For passing test we consider the test is correct or not. For failing tests, we consider two cases either base language errors or OpenACC errors. In addition, we adapt the correctness metric introduced in ~\cite{Godoy_2023} to evaluate the quality of generated tests based on our observations. This metric defines five levels of correctness and proficiency, from 0 (no knowledge) to 1 (expert). We discuss the analysis in Section~\ref{subsec:outputanalysis}. Results from Stage 3 are shown in two tables. Table~\ref{tab:deepseekresults} shows the results of manual analysis of the testsuites generated in the previous stage and Table~\ref{tab:passperlang} shows the results with respect to base language pass percentage.

   These three stages have been enlightening in a way as they have provided tremendous insights for determining how we can improve the overall development process. We discuss this further in the Section~\ref{sec:discussion}.

\section{Results}
\label{sec:results}

In this section, we summarize the results from our evaluation of LLMs and their capabilities in OpenACC test generation. 

\subsection{LLMs}
    To ensure a consistent programming environment to evaluate the performance of selected LLMs and methods, we utilize Anaconda to create a virtual environment and install necessary packages \cite{anaconda}. To locally host open-source LLMs, we use the Llama library for Codellama-34b-Instruct with published inference code \cite{codellamagit} and Huggingface transformers \cite{wolf2020huggingfaces} for Phind-Codellama-34b-v2~\cite{Phind_2023} and Deepseek-Coder-33b-Instruct~\cite{guo2024deepseekcoder}. We use Langchain \cite{langchain} for text-splitting as a preprocessing step for input into the vector database. We use scikit-learn vector store~\cite{sklearn} with Huggingface embeddings through Langchain to create the vector database. For GPT-3.5-Turbo and GPT-4-Turbo we use the OpenAI API for all inference and fine-tuning \cite{openaiapi}. For fine-tuning open-source LLMs, we use Huggingface transformers and the supervised fine-tuning trainer from the Transfer Reinforcement Learning library~\cite{vonwerra2022trl}. For efficient multi-node training we used PyTorch FullyShardedDataParallel with DeepSpeed ZerO-3 Optimization, and PEFT LORA with bf16 precision and flash-attention-2~\cite{dao2023flashattention2, hu2021lora, rajbhandari2020zero, zhao2023pytorch}. To evaluate the generated tests, we create a python script to compile and run tests using NVIDIA-HPC-SDK 23.5, and record the results \cite{nvidia-hpc-sdk}.


\subsection{Experimental Setup}	

    All local inference and fine-tuning for open source  were performed using the National Energy Research Scientific Computing Center's (NERSC) Perlmutter~\cite{perlmutter}, an HPE Cray EX supercomputer equipped with AMD EPYC CPUs and NVIDIA A100 GPUs and an NSF-sponsored HPC Cluster in UDEL, Darwin~\cite{darwin} equipped with AMD EPYC and Intel Platinum CPU nodes and NVIDIA V100 GPUs and one AMD MI100 GPU.

\begin{table*}[!tb]
  \centering
    \begin{tabular}{l|ccccc}
    \toprule
    LLM & True Pass & Pass Correctness & Fail Correctness & Base Lang Error & OpenACC Error \\
    \midrule
    Deepseek-Coder-33b-Instruct & 76 & 0.85 & 0.38 & 47 & 76 \\
    \bottomrule
    \end{tabular}%
  \caption{Shows Stage 3 results. These are the results from manual analysis of Stage 2 generated testsuite. These results specifically focus on tests generated by Deepseek-Coder-33b-Instruct. True Pass indicates the percentage of passing tests that are correct. Pass Correctness and Fail Correctness are the average Correctness score as defined in Section~\ref{sec:stages} for passing and failing tests respectively. Base Language Error and OpenACC Error indicate the percentage of these errors in failing tests. }
  \label{tab:deepseekresults}%
\end{table*}%

\begin{table}[t]
  \centering
    \begin{tabular}{l|ccc}
    \toprule
    LLMs & C & C++ & Fortran \\
    \midrule
    Phind-Codellama-34b-v2 & 42 & 41 & 19 \\
    Codellama-34b-Instruct & 15 & 21 & 4 \\
    Deepseek-Coder-33b-Instruct & \textbf{51} & \textbf{47} & \textbf{47} \\
    GPT-4-Turbo & 50 & 37 & 34 \\
    Fine-tuned GPT-3.5-Turbo & 32 & 3 & 11 \\
    Fine-tuned Phind & 34 & 31 & 15 \\
    Fine-tuned Deepseek & 49 & 44 & 44 \\
    \bottomrule
    \end{tabular}%
    \caption{Table indicates pass percentage per base language for every LLM. The results in bold indicate the highest pass percentage}
  \label{tab:passperlang}%
\end{table}%

\subsection{Codellama-34b-Instruct}
\label{sec:Codellama-34b-Instruct}
We first generate a testsuite in C using Codellama-34b-Instruct using the prompt methods listed in Section~\ref{sec:prompt}
We run each generated testsuite against an OpenACC compiler using a script to capture the number of compile errors, runtime errors, and passing tests. The passing percentage is displayed in Table~\ref{tab:stage1}.

Codellama-34b-Instruct produced the most passing tests using the expressive prompt using RAG with a code template. This is not necessarily indicative of best performance, as the goal of the compiler validation testsuite is not to achieve no errors, but rather to detect existing errors in implementations. Thus, errors are expected to be acceptable outcomes of the testsuite. However, from our previous work~\cite{openaccaaron}, we know that the NVIDIA compilers implement a majority of the specification, thus approaching a similar level of passing tests is a first step.

In Stage 2, as described in Section~\ref{sec:stages}, using the expressive prompt with RAG and template, we generate a testsuite in C, C++, and FORTRAN. 
As indicated earlier in the same section that there are 351 tests that could be generated out of the two chapters of the specification, 47/351 produce passing results by checking for return code ``zero,'' shown in Table~\ref{tab:stage2results}. The bulk of the errors are compile errors. The 66 parsing errors result from the LLM generating a seemingly infinite loop of code such as variable declarations that is larger than the context window or output limit.

We also found that only 4\% Fortran tests in Stage 2, while for C and C++ 15\% and 21\% passed respectively, shown in Table~\ref{tab:passperlang}. This suggests that future training for LLMs for this task should include more Fortran code within the training dataset.

\subsection{Phind-Codellama-34b-v2 and its Fine-tuned version}
From Stage 1 results, we find that Phind-Codellama-34b-v2 produced almost double the number of passing tests using RAG with a code template than RAG with a one-shot example. Additionally, we note more passing tests are produced without using RAG with this model, which is not the case, for example, with Deepseek-Coder-33b-Instruct.
Providing the latest version of the OpenACC specification through RAG is essential to producing up-to-date tests, given that the specification is updated yearly and it is unclear what data relevant to the OpenACC specification LLMs have been pre-trained on. 
Thus, we retain RAG in Stage 2, selecting again an expressive prompt with a template and RAG, as with Codellama-34b-Instruct.

In Stage 2, Phind-Codellama-34b-v2 produced 119/351 passing tests in the generated testsuite by checking for return code ``zero''. 
The majority of errors are compile errors rather than runtime or parsing errors. We find that in comparison to Codellama-34b-Instruct, Phind-Codellama-34b-v2 produced less incomplete tests, shown as parsing errors in Table \ref{tab:stage2results}. In addition, Phind followed a similar trend to Codellama in passing percentage per language, with Fortran lagging behind C and C++. Another observation from Stage 2 is the decrease in passing rate after fine-tuning Phind-Codellama-34b-v2, which we plan to explore further in future works. 



\subsection{Deepseek-Coder-33b-Instruct and its Fine-tuned version}
In Stage 1 we find that Deepseek-Coder-33b-Instruct produced the most passing tests with 3 out of 5 of the prompting methods, compared with the other LLMs. The prompts with a code template and RAG produced 56\% passing tests, more than any other LLM or setup. The performance across prompting methods in Stage 1 with this model is also consistently around 50\%. In comparison to Codellama-34b-Instruct, this LLM is far outperforming in Stage 1.

In Stage 2 Deepseek-Coder-33b-Instruct produced the most passing tests of any LLM: 170/351. It also performed consistently across C++ and Fortran with a slight increase in passes for C tests. We discuss the manual analysis of this testsuite in Section \ref{subsec:outputanalysis}. We also note that pass percentage slightly decreased for this LLM after fine-tuning, as with Phind-Codellama-34b-v2.

\vspace{10pt}

\subsection{GPT-3.5-Turbo and its Fine-tuned version}

We compare the performance of OpenAI's GPT-3.5-Turbo using the same five prompting methods as listed in Subsection~\ref{sec:prompt}. We also evaluate the performance of fine-tuned version of GPT-3.5-Turbo as mentioned in Subsection~\ref{sec:finetuning}. 
Table~\ref{tab:stage1} shows that GPT-3.5-Turbo with no fine-tuning using an expressive prompt with RAG and template produced more passing tests than other methods, by checking for return code ``zero''.

We fine-tune GPT-3.5-Turbo using the dataset comprising of 1335 training examples for 3 epochs through the OpenAI API.
The generated tests using a simple prompt followed the testing format shown in fine-tuning examples without an example, so the results are initially promising.   
We find in the Stage 1 results that this model produced 27\% passing tests, slightly better than before fine-tuning: 23\% with the same prompt. We include this fine-tuned model in the final analysis for a comparison with LLMs not fine-tuned on an OpenACC validation test dataset.

In Stage 2 we find that the fine-tuned GPT-3.5-Turbo produced 85 passing tests by checking for return code ``zero'', which is not a majority, but does outperform Codellama-34b-Instruct. The fine-tuned GPT-3.5-Turbo also produced a higher percentage of passing tests in comparison to the base version of LLM in Stage 1, across all methods. We think that the results may have improved for GPT-3.5-Turbo, while not for open source LLMs, due to the training implementation done by OpenAI in comparison to parameter-efficient methods we used in fine-tuning open source LLMs. Only 11\% of generated Fortran tests passed while 32\% and 30\% pass in C and C, a similar trend to Codellama-34b-Instruct and Phind-Codellama-34b-v2.


\subsection{GPT-4-Turbo}

We evaluate the performance of OpenAI's GPT-4-Turbo in Stage 1 with the 5 prompt methods listed in Section~\ref{sec:prompt}. 
GPT-4-Turbo produced the most passing tests using either only a oneshot example, or an expressive prompt with RAG and a code template.
In this case, one-shot prompting is not outperformed by a code template, as we see with other LLMs. In this stage, we see that oneshot prompts and expressive prompts using RAG with code template produce the most passing tests, and we include the latter in Stage 2 to provide a consistent comparison among all LLMs.

GPT-4-Turbo with expressive prompt using RAG with code template for Stage 2 testsuite generation produced 142 passing tests out of 351, outperformed in this respect only by Deepseek-Coder-33b-Instruct. It also did not succumb to many parsing errors. The passing percentage for C tests is 50\% while C++ and Fortran tests it is lower, at 37\% and 34\%, following a similar trend to other LLMs.
The results are tabulated in Tables~\ref{tab:stage2results} and \ref{tab:passperlang}.

\begin{figure}[!t]
\centering
\begin{lstlisting}[frame=single, numbers=left, language=Fortran]
!T1:serial construct,V:2.7-2.3
      LOGICAL FUNCTION test1()
        USE OPENACC
        IMPLICIT NONE
        INCLUDE "acc_testsuite.Fh"
        
        INTEGER :: errors = 0
        INTEGER :: a(10), b(10), c(10)
        INTEGER :: i

        ! Initialize arrays
        DO i = 1, 10
          a(i) = i
          b(i) = i * 2
          c(i) = 0
        END DO

        !$acc parallel loop
        DO i = 1, 10
          c(i) = a(i) + b(i)
        END DO

        !$acc serial
        DO i = 1, 10
          IF (c(i) .ne. a(i) + b(i)) THEN
            errors = errors + 1
          END IF
        END DO

        IF (errors .eq. 0) THEN
          test1 = .FALSE.
        ELSE
          test1 = .TRUE.
        END IF
      END
\end{lstlisting}
\caption{Failed test for \textit{serial} in Fortran generated using Deepseek-Coder-33b-Instruct. This test results in a compiler error because the \textit{serial} construct requires a closing directive when used on a loop.}
\label{fig:failedtest2}
\end{figure}

\subsection{Deepseek-Coder-33b-Instruct Output Analysis}
\label{subsec:outputanalysis}

To systematically analyze the tests generated by Deepseek-Coder-33b-Instruct, we manually evaluate a representative subset of 25 passing tests and 25 failing tests. 
To evaluate why tests are failing, we consider two cases: incorrect implementation of OpenACC, or failure despite a correct OpenACC implementation. 
Incorrect implementation of OpenACC code means that a feature of OpenACC is used incorrectly, thus causing a failure through compile error, runtime error, or a non-zero return value. 
A failure despite a correct OpenACC implementation means that the OpenACC logic could be correct, but there is an error in the code somewhere else. This could be a syntax error, use of a undefined function, incorrect testing logic, or an unimplemented feature or anything else that may cause a compile error, runtime error, or non-zero return value. 
An example of such a scenario could be a test that randomly fills an array with numbers, then squares all elements within an OpenACC construct, but then forgets to square the elements of the expected output.
In addition to assessing passing tests and causes of failure in failing tests, we analyze the quality of each generated test as defined by the correctness metric described in Section \ref{sec:stages}.

We find that 76\% of the selected passing tests are properly testing the target feature, while the rest are not, shown in Table~\ref{tab:deepseekresults}. In some cases where the test is incorrect, a different feature than the intended one is being tested, or the correct feature is being tested improperly. In analysis of failing tests we find that 47\% have an error in the base language, and 76\% have an error in OpenACC implementation.

Throughout this analysis, we find that the majority of passing tests are correctly implementing OpenACC. The presence of false passing tests  reaffirms that human evaluation is necessary for this task. Complete automation is not yet plausible with the techniques described here. In failing tests, we find a variety of errors consisting of a somewhat even distribution of errors due to incorrect OpenACC implementation, base language errors, and testing logic.

In Figure \ref{fig:failedtest2}, a Fortran test for the serial construct generated by Deepseek-Coder-33b-Instruct is shown.  The test logic is sound, but the usage of OpenACC in Fortran is incorrect. The \textit{serial construct} requires a closing directive, \textit{!\$acc end serial}, when used with a for-loop. Therefore, this test fails. In the evaluation of the output of tests throughout this research, we saw many cases where the testing logic is sound, and the test is close to being correct, but it misuses an OpenACC feature in some way. This insight tells us that the LLM needs improved knowledge of the syntax and specification of OpenACC whether through training or within the prompt. 

\begin{figure}[h]
\centering
\begin{lstlisting}[frame=single, numbers=left, language=Python]
int test1(){
    int err = 0;

    #pragma acc parallel num_gangs(2)
    {
        int gang_id = acc_get_gang_num();
        printf("Hello from gang %d\n", gang_id);
    }

    return err;
}
#endif
\end{lstlisting}
\caption{Failed test for \textit{parallel num gangs} clause in C generated using Deepseek-Coder-33b-Instruct. This test results in a compiler error because the routine \textit{acc get gang num} does not exist; the LLM hallucinated. Moreover, the test is not validating the any computation or effect of the \textit{num gangs clause.}}
\label{fig:failedtest1}
\end{figure}

Figure \ref{fig:failedtest1} shows the logic of a test generated for the parallel construct num gangs clause with Deepseek-Coder-33b-Instruct in C. It demonstrates multiple common issues: First, the routine used in the code, \textit{acc get gang num}, is not a part of the OpenACC specification and is a hallucination of the LLM, causing the test to fail. Next, the test does not perform any real computation with the feature, nor does it validate the output after using the feature. A proper way to test this feature might involve creating for loop to modify an array of values and specifying the number of gangs to be used in parallelization of this for loop on the accelerator. Then, the modification of the array can be validated by performing the same computation on the CPU, without using OpenACC, and comparing the results.

Some other errors were common among the failed tests. Redundant data movements occurred even in passing tests, a cause for concern and a sign of an incomplete understanding of the OpenACC programming paradigm. Many test failed due to undefined constants such as the random seed, an error would could be remedied by providing compiler feedback to the LLM and giving it a chance to revise the code. Improper usage of other OpenACC features within a test despite proper usage of the target feature caused other tests to fail. Often a failing test of this type would be overly complicated, involving many different features potentially conflating the interpretation of the test for a specific feature. Routine misuse and hallucination of routines was also a common error. 

\subsection{Results Summary}
Drawing a summary out of the above discussed results and tying them to our contribution list from Section~\ref{subsec:motivation}, we infer the following: 
\begin {itemize}
\item Out of the five prompt sets evaluated, we find the best performance with an expressive prompt using RAG and a test template.
\item For this study the fine-tuning dataset enabled LLMs to produce correct format of test without a one-shot example or code template, as expected. However, the test quality was not generally improved,  indicated by low passing rate.
\item Out of the seven LLMs evaluated in Stage 2, we find that Deepseek-Coder-33b-Instruct produces the most passing tests.
\item In the final results analysis, we find that most passing tests implement correct OpenACC logic, while failing tests fail for a variety of reasons.
\end{itemize}



\section{Discussion}
\label{sec:discussion}
    By automating the process of generating functional test cases, we are enabling the developers to not spend their time to write these tests, instead spend their time on designing regression and corner case tests - tests that we have not yet fully explored LLMs for nor are we aware at this point how suitable LLMs would be for those cases. 
    Moreover, validation testsuite generation are not a one and done task. The specification of programming models constantly evolves. So a validation testsuite must be maintained by a team of developers to be consistent with the latest version of the specification.  
    Until recently, natural-language processing techniques were not advanced enough to comprehend a long, complex programming model specification and generate code to validate compiler implementations of the specification. With LLMs, its much more viable. 
    When the specification is updated, an LLM can be used again to generate an up-to-date testsuite. 
    
    We found that our initial expectations of the relative performance of the selected LLMs, based on benchmarks and related work, were confirmed by our evaluation. 
    We did not expect this research in its first iteration, to necessarily implement a perfect testsuite generation pipeline, but rather provide insight to the capabilities of current LLMs and areas for improvement and motivation to adapt ideas to several other testsuite projects. We did not notice major differences between different LLMs output besides the frequency of making mistakes causing failed tests, such as described in Section~\ref{subsec:outputanalysis}.
    The performance of the selected LLMs on the HumanEval benchmark trends similarly to the pass rates and performance we found in OpenACC testsuite generation. Though this is a reassuring result, we plan to create a high-performance computing code generation benchmark especially with a focus on testing for LLMs in the future.

 To discuss the results further, first, we find that the use of a simple code template rather than a one-shot example of test implementation produced more passing tests. The difference between these methods is that the template does not provide any example OpenACC code or testing logic. This is interesting to observe because the full example provides more context. 
  
    A potential cause in this result is the increase in length of the input creating more complexity. It introduces potentially unrelated OpenACC features into the prompt, and increases the number of tokens in the input. The prompt all-together is long: when using RAG, we include in each prompt the corresponding section of the specification for the feature being tested. The prompt itself is expressive, and the one-shot example is a full test. All of this adds up to a long prompt, and additional complexity from a full one-shot example seems to negatively impact the test quality in this research.
    
    Using a simple template reduces the input length and complexity, allowing LLM to focus on the feature at hand, and not the feature in the example. The use of an expressive prompt, detailing the goal of the test and the desired behavior, generally improved performance, despite the increase in input length. We believe that the expressive prompt includes relevant context to the task, while the one-shot example provided information about irrelevant OpenACC features, thus confusing the LLM leading to hallucinations.

    Based on the manually created OpenACC testsuite results, we initially expected higher pass rates for the testsuites generated in this work. 
    The OpenACC V\&V testsuite produces 81.4\% passing tests (1087/1335) while the highest passing rate from Stage 2 using Deepseek is 48\% (170/351), using the same hardware and environment.
    An important distinction here is the number of tests. The manually written testsuite consists of 1335 tests, while the full test generated in this work consist of 351 tests. The manually written testsuite implements a full coverage of the OpenACC specification with the exception of a few Fortran routines added in recent updates. The testsuites generated in this work provide a thorough coverage of the specification by generating tests based on sections within the specification, but they consider far fewer permutations of features and uses of features than the manually written testsuite.
    Thus, number of tests for each feature is not consistent between the tests generated in this work and the manually written suite, so the passing rate is not a fair comparison. Nevertheless, we find through manual analysis that the majority of failing tests fail due to issues in the tests, rather than compiler implementations, suggesting further development is required before putting any of the generated testsuites into production.

    We discuss Deepseek-Coder-33b-Instruct output analysis in Section~\ref{subsec:outputanalysis} and 
    the different scenarios where tests fail.

     We find that the incorrect use of OpenACC features sometimes occurred not due to incorrect use of the feature being tested, but rather due to improper use of other OpenACC features within the test. For example, the generated test might target the copyin clause, but the LLM may implement another clause in the test, and use it incorrectly. 
    A potential solution to this source of error is to implement a RAG method that includes the sections from the specification for all OpenACC features used in each test, not only the single feature being tested. This solution would involve two steps. First, planning of a test. Then, retrieval of all relevant sections, and second, the generation of the test.
    
    Another method that we suggest could improve the performance of this work greatly is to give the LLM multiple attempts at each test. 
    This could involve simultaneous generation of multiple tests, compilation of each and running of each, analysis of the output, and selection of the best test based on the output. Within LLM benchmarking, this method is known as the pass@k metric \cite{chen2021evaluating}, where LLMs are given more than one attempt at completing the task of the benchmark, e.g. code generation. The difficulty in automation compiler validation test generation via brute force generation of many solutions is the lack of a automated metric for evaluating test quality and accuracy: this still requires a human, an expert. Thus, the methods used to train and inference using the LLMs must be developed further for the quality of test generation to improve. 

    Despite the gap in LLM capability to automate validation test generation, we believe it is a handy tool for skilled developers to use. The process of developing a large testsuite and maintaining it through changes to the specification is lengthy and recurrent. If an experienced developer can quickly generate a near-accurate test and fix it, or generate permutations of or add additional test cases to a working test, there are potential benefits to the developers' efficiency. Regardless, looking forward, it is unclear that a LLM-generated validation testsuite can be put into production without oversight from a developer.

\section{Conclusion}
\label{sec:conclusion}
The purpose of this work is to apply LLMs to generation of OpenACC test cases which in turn would be used to validate and verify compiler implementations of the OpenACC specification. As the specification is updated, the pipeline can be run again to generate a new suite of tests that verify the newly added features. Moreover, the improvement of techniques in optimizing large language model performance will likely improve the overall performance of test case generation. This way we aim to ``shift'' where the developers use their valuable time and not ``replace'' their efforts. This project requires human oversight but at the same time also relieves them from writing tests that LLMs can do a good job at. 

This project is meant as a new application of the emerging LLMs in high performance computing. As an immediate and on-going work we have have started to look into optimizing our methods and exploring the adaptability to the OpenMP testsuite. As near future steps we will work with developers of Kokkos, RAJA, Chapel, SYCL among others and explore the usability and expandability of the approach based on their inputs and needs.

The harsh truth about the current state of LLMs is that they can generate false and misleading output at times, despite various efforts to discourage this behavior and we know that LLMs hallucinate. We emphasize the need to exercise caution and conduct human evaluation on all outputs that are used in production. The accuracy of these methods is likely to improve as the community continues to learn and make adjustments.

\section{Acknowledgments}

This research was supported by the National Science Foundation (NSF) under grant no. 1919839, in part through the use of DARWIN computing system: DARWIN - A Resource for Computational and Data-intensive Research at the University of Delaware. This material is also based upon work supported by NSF under grant no. 1814609.

This research also used resources of the Oak Ridge Leadership Computing Facility at the Oak Ridge National Laboratory, which is supported by the Office of Science of the U.S. Department of Energy under Contract No. DE-AC05-00OR22725.

This research also used resources of the National Energy Research Scientific Computing Center (NERSC), a U.S. Department of Energy Office of Science User Facility located at Lawrence Berkeley National Laboratory, operated under Contract No. DE-AC02-05CH11231 using NERSC award.
We are grateful to OpenACC for their support on this project.

\bibliographystyle{elsarticle-num}
\bibliography{main.bib}
    
\end{document}